\documentclass[10pt,twocolumn,letterpaper]{article}

\usepackage{cvpr}

\definecolor{cvprblue}{rgb}{0.21,0.49,0.74}
\usepackage[breaklinks,colorlinks,allcolors=cvprblue]{hyperref}
\usepackage{multirow}

\def\paperID{29443}
\def\confName{CVPR}
\def\confYear{2026}

\title{\textit{YCDa}: \textit{YC}bCr \textit{D}ecoupled \textit{A}ttention for Real-time Realistic Camouflaged Object Detection}

\author{
PeiHuang Zheng,
Yunlong Zhao,
Zheng Cui,
Yang Li*\\
Nanjing University of Aeronautics and Astronautics\\
\{bluce\_8185,zhaoyunlong,cuizheng,liyangnuaa\}@nuaa.edu.cn
}

\begin{document}
\maketitle
\pagestyle{plain}

\begin{abstract}
Human vision exhibits remarkable adaptability in perceiving objects under camouflage. When color cues become unreliable, the visual system instinctively shifts its reliance from chrominance (color) to luminance (brightness and texture), enabling more robust perception in visually confusing environments. Drawing inspiration from this biological mechanism, we propose \textbf{YCDa}, an efficient early-stage feature processing strategy that embeds this “chrominance–luminance decoupling and dynamic attention” principle into modern real-time detectors. Specifically, YCDa separates color and luminance information in the input stage and dynamically allocates attention across channels to amplify discriminative cues while suppressing misleading color noise. The strategy is plug-and-play and can be integrated into existing detectors by simply replacing the first downsampling layer. Extensive experiments on multiple baselines demonstrate that YCDa consistently improves performance with negligible overhead as shown in Fig.~\ref{fig:performance_overview}. Notably, \textbf{YCDa-YOLO12s} achieves a \textbf{112\% improvement in mAP} over the baseline on COD10K-D and sets new state-of-the-art results for real-time camouflaged object detection across COD-D datasets.
\end{abstract}

\noindent\textbf{Code:} \url{https://github.com/hhao659/YCDa}


\begin{figure}[t]
\centering
\includegraphics[width=0.47\textwidth]{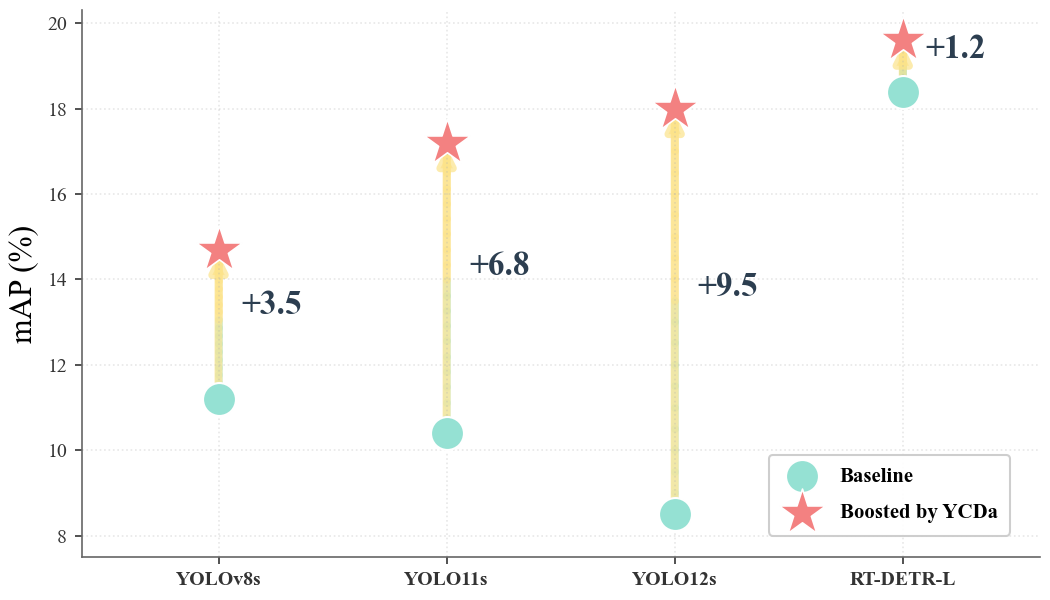}
\caption{\textbf{Improvements over different baseline models.} The results are reported on the COD10K-D test set in mean Average Precision (mAP).}
\label{fig:performance_overview}
\end{figure}

\section{Introduction}
Realistic Camouflaged Object Detection (RCOD) is a recently proposed object detection task designed to address the inefficiency and high annotation costs associated with traditional Camouflaged Object Detection (COD)~\cite{xin2025toward}. RCOD only requires bounding box annotations to localize camouflaged objects and identify their categories, substantially reducing labeling costs while maintaining practical utility. However, unlike segmentation-based COD methods that rely on pixel-level contrast between object and background, RCOD models operate at the region level, where fine-grained cues are often lost, making camouflaged detection significantly more challenging.

Large-scale pre-training and robust semantic extractors have improved object detection accuracy across many domains~\cite{li2022grounded,cheng2024yolo}. Yet, for real-time scenarios such as search-and-rescue~\cite{bachir2024benchmarking} or military surveillance~\cite{zhuang2024military}, existing detectors like YOLO~\cite{redmon2016you,Jocher_Ultralytics_YOLO_2023,yolo11_ultralytics,tian2025yolov12} still struggle to maintain reliable performance under camouflage, where visual cues are intentionally suppressed. These situations demand models capable of robust perception under deceptive visual conditions while retaining efficiency.

\begin{figure*}[t]
\centering
\includegraphics[width=1.0\textwidth]{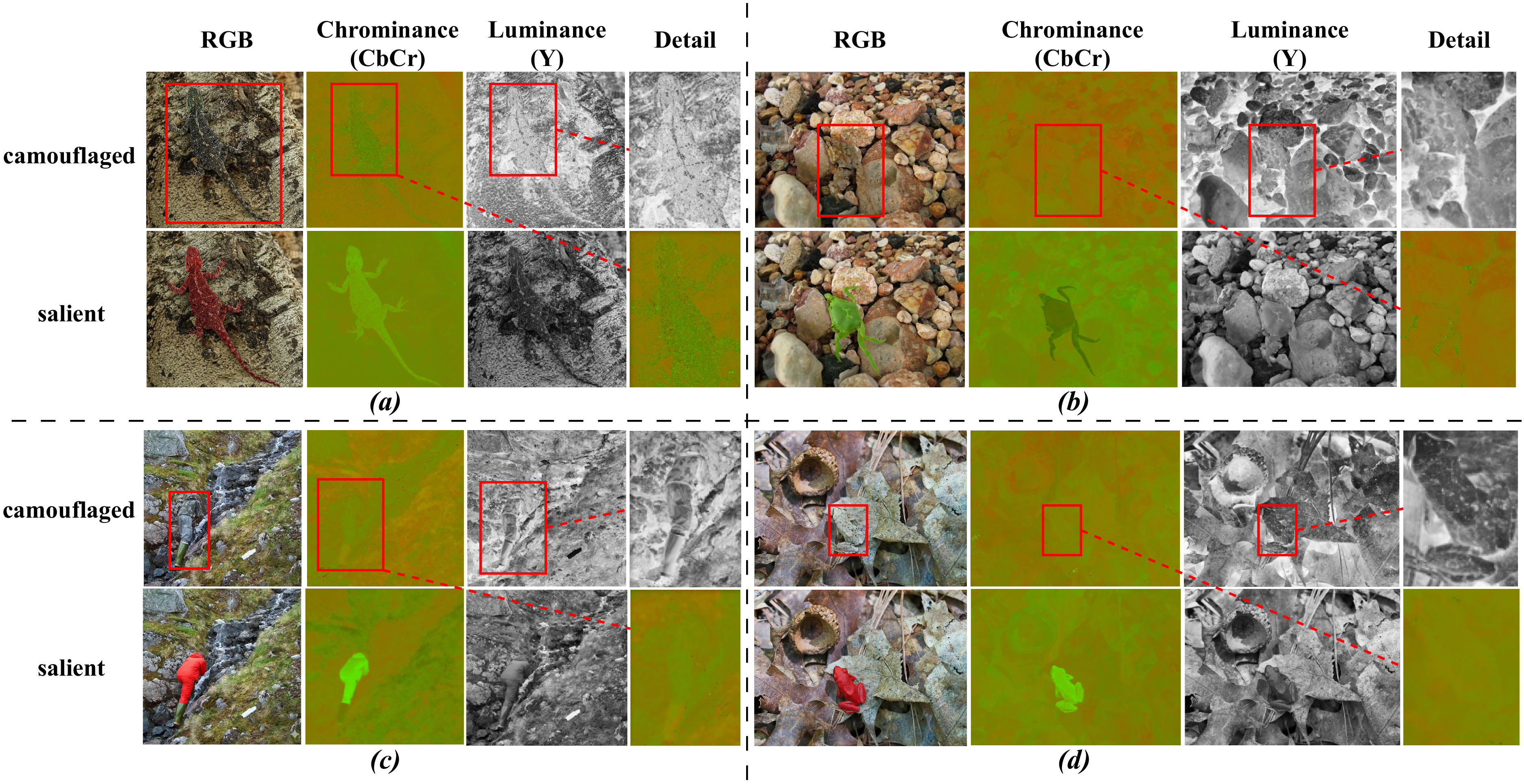}
\caption{\textbf{Comparison of chrominance and luminance channels under different object saliency levels.} Each group contains camouflaged objects (the first column) and salient objects (the second column) generated by modifying only color while preserving original features.}
\label{fig:chroma_luma}
\end{figure*}

Human vision, in contrast, exhibits remarkable adaptability in such camouflaged environments. When color cues become unreliable, the visual system instinctively shifts its focus from chrominance (color) to luminance (brightness and texture) to maintain stable perception~\cite{gegenfurtner2003color,kane2011delays,negishi2021suppression,clery2013interactions}. This mechanism has been observed in both physiological studies and behavioral phenomena, such as reports that color-deficient snipers outperformed trichromatic observers in camouflage detection. The biological explanation is intuitive~\cite{morgan1992dichromats}: when background and target colors are similar, excessive sensitivity to chrominance can be misleading, while luminance and texture provide more robust signals for boundary detection. 

Inspired by this biological principle, we investigate how chrominance and luminance information contribute differently to object discrimination under varying saliency levels. As illustrated in Figure~\ref{fig:chroma_luma}, when objects are salient, chrominance offers strong discriminative cues, whereas luminance cues are less informative. However, as camouflage deepens and color differences vanish, chrominance channels not only lose effectiveness but can even introduce misleading information. In contrast, luminance channels retain textural and contour features essential for object recognition. These observations suggest that models treating chrominance and luminance information equally may overemphasize noisy color signals and neglect valuable luminance cues, resulting in poor camouflage detection performance.

To address this issue, we propose \textbf{YCDa} (\underline{\textbf{Y}}CbCr \underline{\textbf{D}}ecoupled \underline{\textbf{A}}ttention), a biologically inspired, plug-and-play early-stage feature processing strategy that embeds the “chrominance–luminance decoupling and dynamic attention” mechanism into modern real-time detectors. Specifically, YCDa first transforms the input into the YCbCr color space to explicitly separate luminance and chrominance channels, preserving their distinct roles in perception. It then applies a pointwise-convolution-free downsampling module~\cite{zheng2025prnet} to prevent premature channel mixing, and an Information-aware Channel Attention (ICA) module that adaptively adjusts channel weights based on variance statistics—amplifying discriminative information while suppressing misleading chromatic noise. This design not only enhances robustness under camouflage but also maintains the real-time efficiency crucial for deployment.

Our method achieves consistent gains across multiple real-time detectors with negligible computational overhead. For instance, YCDa-YOLO12s achieves an impressive \textbf{112\% mAP improvement} on COD10K-D, establishing new state-of-the-art results on the COD-D benchmark suite. 

Our main contributions are summarized as follows:
\begin{itemize}
    \item We identify and analyze the fundamental disparity between chrominance and luminance cues under varying camouflage conditions, revealing why conventional detectors struggle to suppress color-induced noise and capture effective structural features.
    \item We propose \textbf{YCDa}, a biologically inspired feature processing strategy that integrates chrominance–luminance decoupling and information-aware dynamic attention, allowing detectors to extract robust features under different saliency levels.
    \item We introduce a lightweight Information-aware Channel Attention (ICA) module that leverages variance information to dynamically reweight feature channels, improving discriminative focus with minimal cost.
    \item Extensive experiments on COD-D benchmarks demonstrate that YCDa delivers superior detection accuracy and real-time performance, setting new state-of-the-art results for camouflaged object detection.
\end{itemize}

\begin{figure*}[t]
\centering
\includegraphics[width=1.0\textwidth]{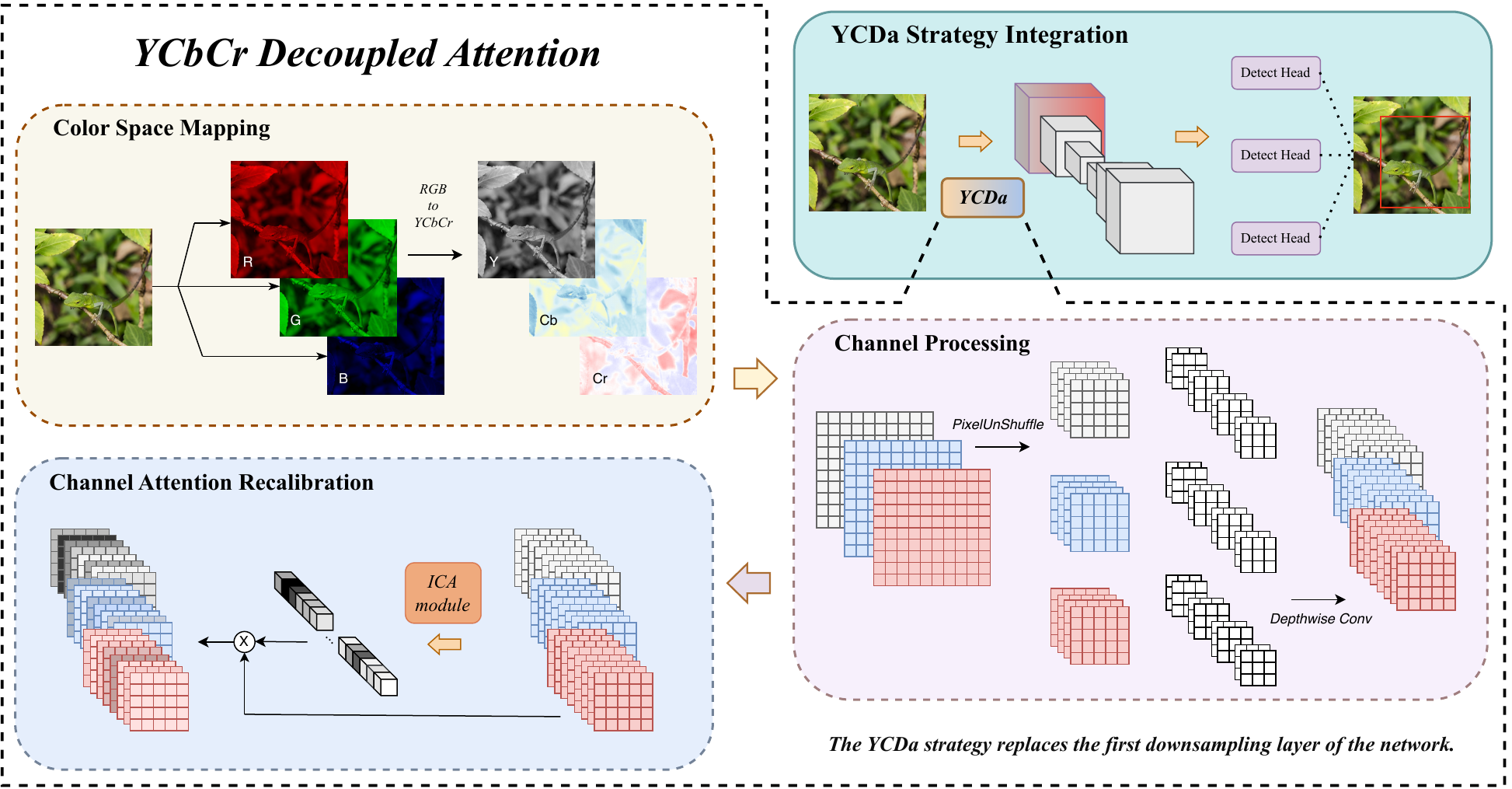}
\caption{\textbf{Overview of YCbCr Decoupled Attention strategy.} The input image first undergoes color space transformation, then uses point-wise-free ESSamp for downsampling and preliminary feature processing, followed by the ICA module to allocate attention across different chrominance and luminance information channels. The top-left corner shows the YCDa-Enhanced Detection Network.}
\label{fig:ycda_overview}
\end{figure*}

\section{Related Works}

\subsection{Realistic Camouflaged Object Detection}
Realistic Camouflaged Object Detection (RCOD) was proposed to detect camouflaged objects using conventional object detection approaches~\cite{xin2025toward}. Compared to previous segmentation-based methods, RCOD offers higher efficiency and lower annotation costs for tasks that only require precise object localization. However, this poses a formidable challenge for detection models that omit pixel-wise analysis. While large-scale vision models with extensive pretraining can effectively perceive objects, they are difficult to deploy in real-time edge scenarios such as search-and-rescue operations and military reconnaissance. Therefore, enhancing the perception capability of real-time object detectors for camouflaged objects while maintaining efficiency has become a critical issue in current RCOD research.

\subsection{Color Space}
Beyond the standard RGB color space, research in image restoration tasks has explored alternative color spaces, such as YCbCr~\cite{fang2025guided,fang2025color}, HSV~\cite{lyu2024mcpnet}, and HVI~\cite{yan2025hvi}. Among these, LCDiff~\cite{fang2025guided} observed that stable patterns in the YCbCr color space can effectively achieve unified removal of weather-induced degradation. By decomposing images into luminance and chrominance components to decouple degraded visual content, independent restoration and unified weather degradation removal are realized. Inspired by these works, we conduct an in-depth exploration of how objects with varying saliency levels change in the YCbCr color space, and propose YCDa to efficiently exploit discriminative information.

\subsection{Attention Mechanisms}
Attention mechanisms have been widely applied in object detection tasks to enhance feature representation capabilities. SENet~\cite{hu2018squeeze} captures inter-channel dependencies through global average pooling, achieving a lightweight channel attention mechanism. CBAM~\cite{woo2018cbam} introduces spatial attention on this basis, refining feature maps through sequential inference of channel and spatial attention. FCA~\cite{qin2021fcanet} captures global contextual information through frequency-domain analysis, enhancing the perception capability of channel attention. However, these methods primarily focus on feature response intensity or frequency-domain information, and struggle to perceive information value differences across channels. In camouflaged object detection tasks, the contributions of chrominance and luminance channels to detection vary significantly under different saliency levels. Traditional attention mechanisms cannot effectively distinguish valuable discriminative information from noise interference, resulting in limited model performance.

\section{Methodology}
In this section, we introduce our proposed YCDa strategy and the Information-aware Channel Attention (ICA) module in detail.

\subsection{YCbCr Decoupled Attention}

Based on our previous analysis, chrominance and luminance channels exhibit distinct contributions to model understanding for objects with varying degrees of saliency. Therefore, to fully exploit valuable discriminative information while suppressing noise interference in the image, we propose an efficient early-stage feature processing strategy called YCDa (YCbCr Decoupled Attention). Figure~\ref{fig:ycda_overview} presents an overview of our method, illustrating the main steps of YCDa: color space transformation, channel processing, channel attention allocation, and the overall architecture using this approach.

\subsubsection{Color Space Transformation.}
We first transform the input RGB image $\mathbf{I} \in \mathbb{R}^{H \times W \times 3}$ to YCbCr color space to explicitly decouple luminance and chrominance information:
\begin{equation}
\mathbf{I}_{YCbCr} = \mathcal{T}_{RGB \to YCbCr}(\mathbf{I})
\label{eq:color_transform}
\end{equation}
where $\mathcal{T}_{RGB \to YCbCr}$ denotes the standard RGB-to-YCbCr conversion, yielding luminance channel $Y$ and chrominance channels $C_b$, $C_r$. This decoupling enables independent processing of color and brightness information.

\subsubsection{Channel-Independent Downsampling.}
To preserve the decoupled channel characteristics while extracting spatial features, we employ a pointwise-convolution-free downsampling strategy inspired by ESSamp~\cite{zheng2025prnet}:
\begin{equation}
\mathbf{F} = \phi(\mathcal{D}(\mathbf{I}_{YCbCr}))
\label{eq:downsample}
\end{equation}
where $\mathcal{D}(\cdot)$ represents PixelUnshuffle that rearranges spatial information into channel dimension, and $\phi(\cdot)$ denotes depthwise convolution that processes each channel independently. Specifically, we assign two convolutional kernels per channel to extract richer features while maintaining channel isolation. This design avoids early feature aliasing and preserves the semantic independence of chrominance and luminance information.

\begin{figure}[t]
\centering
\includegraphics[width=0.48\textwidth]{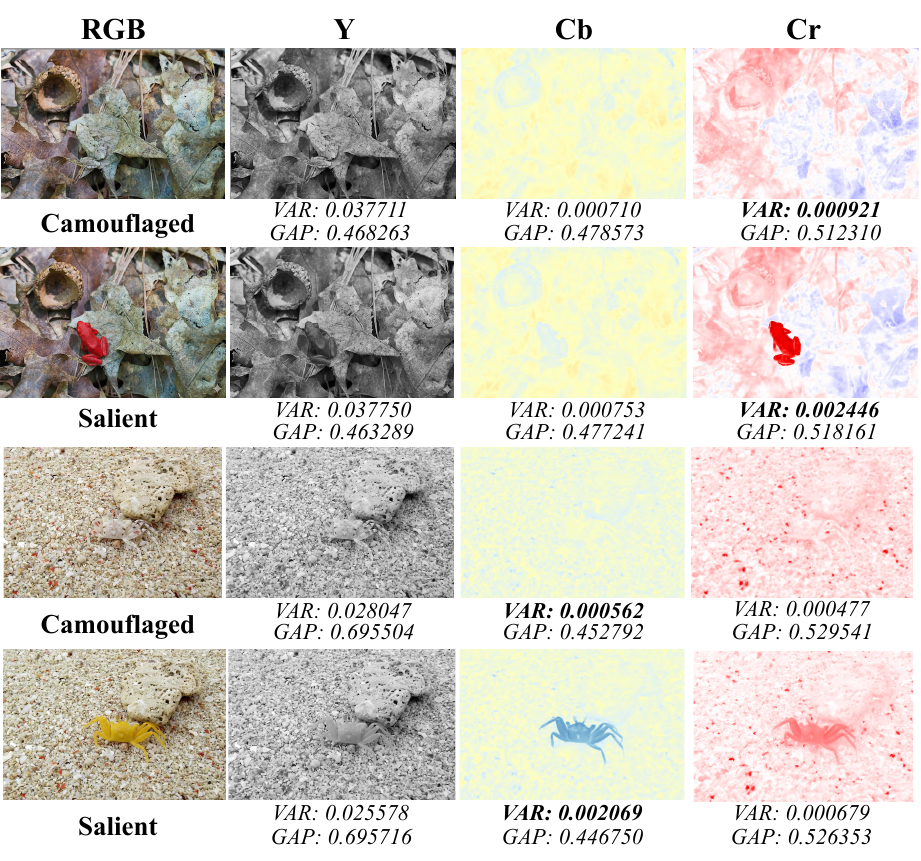}
\caption{\textbf{Comparison of saliency differences, GAP, and VAR across three YCbCr channels.} The visualization demonstrates that variance effectively captures discriminative information value variations across different object saliency levels.}
\label{fig:variance_analysis}
\end{figure}

\subsubsection{Attention Allocation.}
Finally, we apply our proposed ICA module to dynamically weight channel importance based on their discriminative value:
\begin{equation}
\mathbf{F}_{out} = \text{ICA}(\mathbf{F}) \odot \mathbf{F}
\label{eq:attention}
\end{equation}
where $\odot$ denotes channel-wise multiplication, and ICA generates adaptive attention weights for each channel.

YCDa is a plug-and-play efficient method that enhances image understanding, amplifies discriminative information, and suppresses interference by simply replacing the first downsampling layer. Compared to conventional approaches, the additional computational overhead of color space transformation and channel attention allocation in our strategy is negligible. The point-wise-free ESSamp downsampling incurs minimal feature processing cost. More importantly, it reduces the high-resolution output from 64 channels to 24 channels, significantly decreasing the computational load when the subsequent strided convolution performs downsampling. This makes the model perform better while running faster.

\subsection{Information-aware Channel Attention}

Traditional channel attention mechanisms, such as SENet~\cite{hu2018squeeze}, utilize global average pooling to weight attention allocation based on the overall response intensity of each channel. However, this approach focusing on overall response struggles to perceive channel information variation and correctly amplify valuable features while suppressing interference.

Variance is a statistical metric measuring data dispersion. Zero variance indicates identical data values, while larger variance indicates more dramatic data fluctuation. This property aligns perfectly with our assessment of discriminative information value in images. When objects become camouflaged, they typically blend with the environment showing balanced pixel differences; conversely, when targets are salient, pixel differences become pronounced.

\begin{equation}
\text\quad \bar{x} = \frac{1}{HW}\sum_{i=1}^{H}\sum_{j=1}^{W}x_{ij}
\end{equation}
\begin{equation}
\text{Var}(X) = \frac{1}{HW}\sum_{i=1}^{H}\sum_{j=1}^{W}(x_{ij} - \bar{x})^2
\end{equation}
where $H$ and $W$ denote the height and width of the feature map, respectively.

As shown in Figure~\ref{fig:variance_analysis}, we observe that under different object saliency levels, GAP shows almost no variation, the variance of the luminance Y channel changes minimally, while the variance of chrominance Cb and Cr channels exhibits significant differences. Specifically, when the Y channel variance approaches that of the chrominance channels, the target image is more likely to be a camouflaged object or background. In this case, texture and contour information in the Y channel possess greater discriminative value, while chrominance channels tend to introduce noise interference. Conversely, when chrominance channel variance approaches Y channel variance, the image tends toward saliency, with chrominance channels containing more easily distinguishable discriminative information and luminance channel information serving as auxiliary discrimination. 

\begin{figure}[t]
\centering
\includegraphics[width=0.47\textwidth]{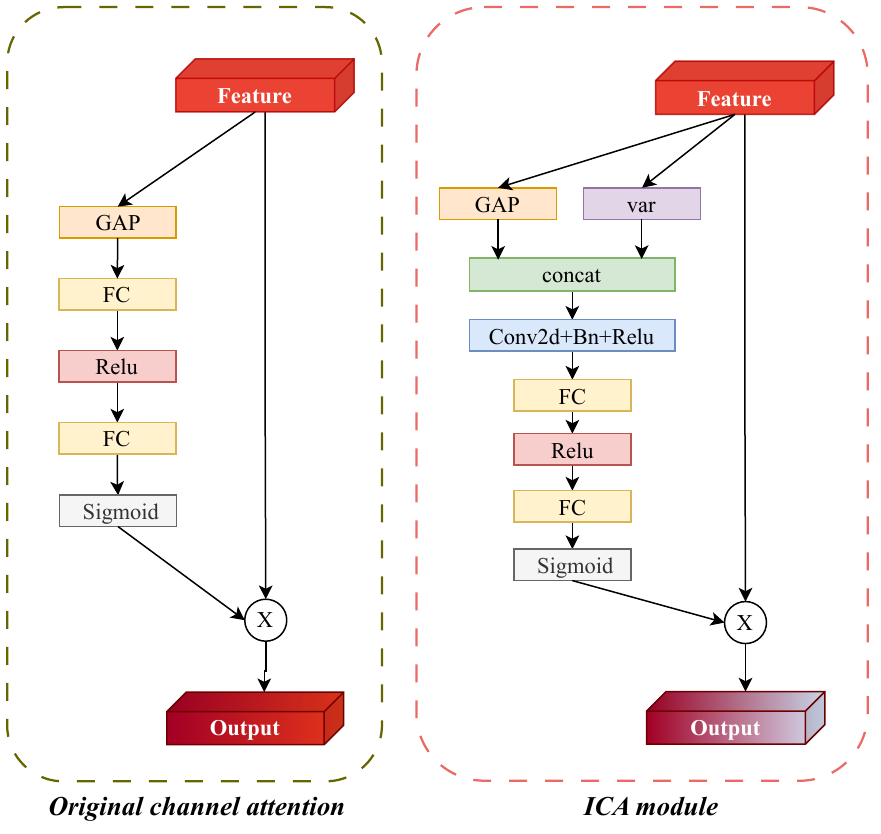}
\caption{\textbf{Architecture of the Information-aware Channel Attention (ICA) module.} The module integrates both global average pooling and variance information to perceive channel-wise information differences, enabling more precise attention allocation.}
\label{fig:ica_module}
\end{figure}

To enable the model to perceive channel information dynamically, we integrate both GAP and variance statistics. First, we compute and fuse the dual statistics:
\begin{equation}
\mathbf{z} = \mathbf{W}_{fuse} \cdot [\text{GAP}(\mathbf{F}); \text{Var}(\mathbf{F})] \in \mathbb{R}^{C}
\label{eq:ica_fusion}
\end{equation}
where $[\cdot; \cdot]$ denotes concatenation, and $\mathbf{W}_{fuse}$ is a $1 \times 1$ convolution that fuses the dual statistics (global average pooling and variance) into a unified representation. This fusion operation extracts complementary information from both response intensity and dispersion patterns.

Subsequently, we apply a standard two-layer MLP with bottleneck structure to generate attention weights:
\begin{equation}
\boldsymbol{\alpha} = \sigma(\mathbf{W}_2 \delta(\mathbf{W}_1 \mathbf{z}))
\label{eq:ica_excitation}
\end{equation}
where $\mathbf{W}_1 \in \mathbb{R}^{\frac{C}{r} \times C}$ and $\mathbf{W}_2 \in \mathbb{R}^{C \times \frac{C}{r}}$ are fully connected layers with reduction ratio $r=4$, $\delta$ denotes ReLU activation, and $\sigma$ represents sigmoid activation. The resulting attention weights $\boldsymbol{\alpha} \in \mathbb{R}^{C}$ are then applied to modulate the input features via channel-wise multiplication.

The architecture of ICA is illustrated in Figure~\ref{fig:ica_module}. By perceiving variance differences across channels, ICA enables the model to distinguish valuable discriminative features from noise interference, dynamically amplifying informative channels while suppressing noisy ones based on object saliency.

\section{Experiments}
\subsection{Implementation Details}

To validate the effectiveness of our proposed strategy, we conduct extensive experiments on the COD-D dataset for realistic camouflaged object detection. COD-D, manually annotated by Huazhong University of Science and Technology, is derived from three existing COD segmentation datasets (COD10K\cite{fan2021concealed}, NC4K\cite{lv2021simultaneously}, and CAMO\cite{le2019anabranch}) with detection labels. It comprises three subsets: COD10K-D, NC4K-D, and CAMO-D, serving as a benchmark for RCOD tasks. We report results using standardized metrics, including mean Average Precision (mAP), AP@0.5, and AP@0.75 across different IoU thresholds.

\begin{table}[ht!]
\centering
\caption{\textbf{Statistics of the three COD-D datasets.}}
\label{tab:dataset_stats}
\resizebox{\columnwidth}{!}{
\begin{tabular}{lcccc}
\toprule
Dataset & Classes & Train Images & Test Images & Background \\
\midrule
COD10K-D & 68 & 6000 & 4000 & \checkmark \\
NC4K-D & 37 & 2863 & 1227 & $\times$ \\
CAMO-D & 43 & 744 & 497 & $\times$ \\
\bottomrule
\end{tabular}
}
\end{table}

\begin{table*}[ht!]
\centering
\caption{\textbf{Performance comparison between YCDa-enhanced models and baselines on COD10K-D, NC4K-D, and CAMO-D datasets.} \textbf{Bold} indicates the best-performing method}
\label{tab:main_results}
\begin{tabular}{lccccccccc}
\toprule
\multirow{2}{*}{Methods} & \multicolumn{3}{c}{COD10K-D} & \multicolumn{3}{c}{NC4K-D} & \multicolumn{3}{c}{CAMO-D} \\
\cmidrule(lr){2-4} \cmidrule(lr){5-7} \cmidrule(lr){8-10}
& mAP & AP50 & AP75 & mAP & AP50 & AP75 & mAP & AP50 & AP75 \\
\midrule
RT-DETR-L\cite{zhao2024detrs} & 18.4 & 26.6 & 19.3 & \textbf{35.9} & \textbf{50.8} & \textbf{36.9} & \textbf{29.3} & \textbf{39.5} & \textbf{31.2} \\
RT-DETR-L + YCDa & \textbf{19.6} & \textbf{28.3} & \textbf{20.7} & 34.1 & 48.7 & 34.0 & 22.5 & 30.1 & 23.1 \\
\midrule
YOLOv8s\cite{Jocher_Ultralytics_YOLO_2023} & 11.2 & 19.9 & 10.9 & 29.0 & 45.1 & 29.9 & 20.5 & 31.1 & 21.2 \\
YOLOv8s + YCDa & \textbf{14.7} & \textbf{24.4} & \textbf{15.1} & \textbf{32.0} & \textbf{48.6} & \textbf{32.8} & \textbf{21.6} & \textbf{31.7} & \textbf{21.5} \\
\midrule
YOLO11s\cite{yolo11_ultralytics} & 10.4 & 17.5 & 10.5 & 27.0 & 40.5 & 28.3 & 21.4 & 31.3 & 21.7 \\
YOLO11s + YCDa & \textbf{17.2} & \textbf{26.3} & \textbf{18.3} & \textbf{31.9} & \textbf{47.0} & \textbf{33.5} & \textbf{25.5} & \textbf{36.1} & \textbf{26.3} \\
\midrule
YOLO12s\cite{tian2025yolov12} & 8.5 & 14.9 & 7.9 & 26.0 & 38.7 & 27.6 & 20.4 & 30.8 & 19.1 \\
YOLO12s + YCDa & \textbf{18.0} & \textbf{28.7} & \textbf{18.2} & \textbf{33.7} & \textbf{48.6} & \textbf{36.6} & \textbf{26.0} & \textbf{35.3} & \textbf{27.7} \\
\bottomrule
\end{tabular}
\end{table*}

\begin{figure*}[ht!]
\centering
\includegraphics[width=1.0\textwidth]{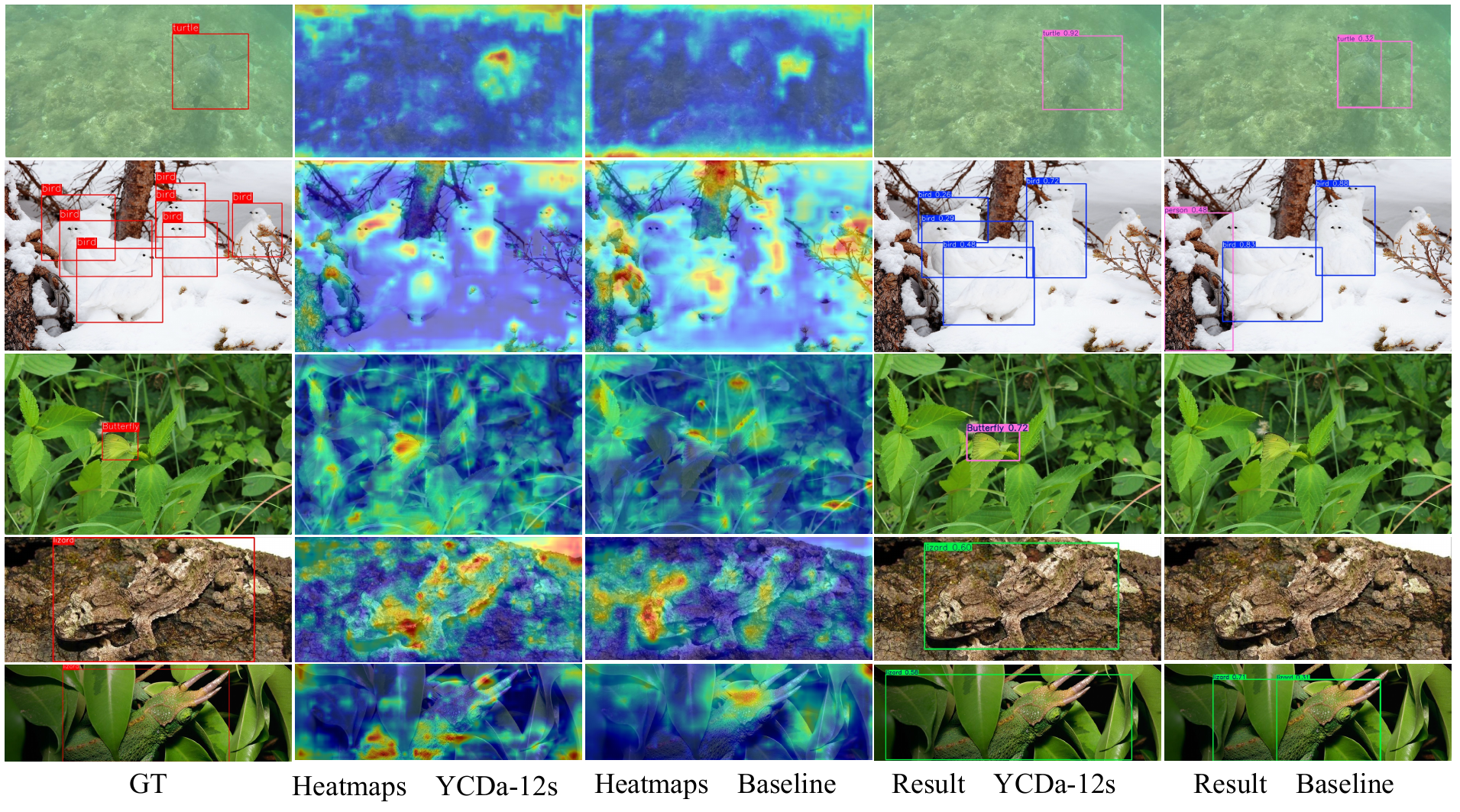}
\caption{\textbf{Qualitative comparison between the baseline YOLO12s~\cite{tian2025yolov12} and YCDa-YOLO12s on COD-D datasets.} The highlighted areas indicate the regions most attended by the network.}
\label{fig:visualization}
\end{figure*}

All experiments are conducted on an NVIDIA GeForce RTX 3090 GPU, except that the inference speed is test on a single RTX 5090 GPU. For CNN-based models using the YCDa strategy, we first perform pretraining on the COCO dataset\cite{lin2014microsoft} for 60 epochs with a batch size of 32. The Transformer-based RT-DETR-L requires longer pretraining, which we set to 250 epochs with a batch size of 32 and an AdamW\cite{loshchilov2017decoupled} optimizer with an initial learning rate of 3e-4. For baseline models, we utilize official pretrained weights from Ultralytics\cite{ultralytics}, maintaining consistent fine-tuning strategies across all models.

Specifically, we employ a Stochastic Gradient Descent (SGD)\cite{robbins1951} optimizer with momentum of 0.937, training for 300 epochs with a batch size of 16, patience of 50, weight decay of 0.0005, and an initial learning rate of 0.01.

\begin{table*}[ht!]
\centering
\caption{\textbf{Real-time performance comparison on COD10K-D.} YCDa-YOLO12s achieves significant accuracy gains with minimal additional cost.}
\label{tab:realtime}
\begin{tabular}{lcccccc}
\toprule
Method & mAP & AP$_{50}$ & AP$_{75}$ & Params (M) & FLOPs (G) & FPS \\
\midrule
YOLO12s~\cite{tian2025yolov12} & 8.5 & 14.9 & 7.9 & 21.69 & 9.28 & \textbf{138.1} \\
YCDa-YOLO12s & \textbf{18.0} & \textbf{28.7} & \textbf{18.2} & \textbf{21.32} & \textbf{9.28} & 131.4 \\
\midrule
Improvement & +112\% & +93\% & +130\% & -1.7\% & 0\% & -4.9\% \\
\bottomrule
\end{tabular}
\end{table*}

\begin{table*}[ht!]
\centering
\caption{\textbf{Ablation study of YCDa components.} YC: YCbCr color space conversion; ES: ESSamp module; ICA: Information-aware Channel Attention.}
\label{tab:ablation}
\begin{tabular}{lccccccccc}
\toprule
\multirow{2}{*}{Methods} & \multicolumn{3}{c}{COD10K-D} & \multicolumn{3}{c}{NC4K-D} & \multicolumn{3}{c}{CAMO-D} \\
\cmidrule(lr){2-4} \cmidrule(lr){5-7} \cmidrule(lr){8-10}
& mAP & AP$_{50}$ & AP$_{75}$ & mAP & AP$_{50}$ & AP$_{75}$ & mAP & AP$_{50}$ & AP$_{75}$ \\
\midrule
Baseline & 8.5 & 14.9 & 7.9 & 26.0 & 38.7 & 27.6 & 20.4 & 30.8 & 19.1 \\
YC & 16.3 & 26.0 & 16.2 & \textbf{32.6} & 48.6 & 34.1 & 23.1 & 34.5 & 24.0 \\
ICA & 16.6 & 25.8 & 17.4 & 32.4 & 46.9 & 33.4 & 20.7 & 29.8 & 21.8 \\
YC + ICA & \textbf{18.3} & 28.5 & 18.3 & 32.4 & 46.2 & 33.8 & 24.5 & 34.1 & 25.4 \\
YC + ES + ICA & 18.0 & \textbf{28.7} & \textbf{18.3} & \textbf{33.7} & \textbf{48.6} & \textbf{36.6} & \textbf{26.0} & \textbf{35.3} & \textbf{27.7} \\
\bottomrule
\end{tabular}
\end{table*}

\begin{table*}[ht!]
\centering
\caption{\textbf{Comparison of different attention mechanisms in YCDa.} SE uses global average pooling; varOnly uses only global variance; all methods use a reduction ratio of 4.}
\label{tab:attention_comparison}
\begin{tabular}{lccccccccc}
\toprule
\multirow{2}{*}{Methods} & \multicolumn{3}{c}{COD10K-D} & \multicolumn{3}{c}{NC4K-D} & \multicolumn{3}{c}{CAMO-D} \\
\cmidrule(lr){2-4} \cmidrule(lr){5-7} \cmidrule(lr){8-10}
& mAP & AP$_{50}$ & AP$_{75}$ & mAP & AP$_{50}$ & AP$_{75}$ & mAP & AP$_{50}$ & AP$_{75}$ \\
\midrule
12s-YCDa-SE & 18.0 & 28.2 & \textbf{18.6} & 32.2 & 46.8 & 33.6 & 21.7 & 33.4 & 22.8 \\
12s-YCDa-varOnly & 17.1 & 27.5 & 16.6 & 32.6 & 47.7 & 33.9 & 22.3 & 32.4 & 22.3 \\
12s-YCDa-CBAM & 17.1 & 27.6 & 17.5 & 31.5 & 46.0 & 33.8 & 20.6 & 32.2 & 19.9 \\
12s-YCDa-FCA & 17.3 & 27.5 & 17.6 & 32.9 & 46.7 & 35.5 & 22.3 & 31.8 & 23.0 \\
12s-YCDa-ICA (ours) & \textbf{18.0} & \textbf{28.7} & 18.3 & \textbf{33.7} & \textbf{48.6} & \textbf{36.6} & \textbf{26.0} & \textbf{35.3} & \textbf{27.7} \\
\bottomrule
\end{tabular}
\end{table*}

\subsection{Results on RCOD Tasks}

To evaluate the effectiveness of the proposed YCDa strategy, we conduct extensive experiments on mainstream real-time detectors across the COD10K-D, NC4K-D, and CAMO-D datasets. As shown in Table~\ref{tab:main_results}, models equipped with YCDa consistently achieve substantial performance gains across all CNN-based baselines.

On the most challenging COD10K-D dataset — characterized by large scale, 68 diverse categories, severe long-tail distribution, and large intra-class variation—baseline detectors exhibit a counterintuitive phenomenon: as the model capacity increases (e.g., from YOLOv8s to YOLO12s~\cite{Jocher_Ultralytics_YOLO_2023,yolo11_ultralytics,tian2025yolov12}), detection accuracy decreases. We attribute this to the models’ over-reliance on color cues, which, though effective in general detection, introduce noise interference in camouflaged scenes. In contrast, YCDa effectively mitigates this bias by emphasizing luminance and texture cues. Specifically, YCDa-YOLO12s~\cite{tian2025yolov12} achieves 18.0\% mAP, representing a remarkable 112\% improvement over the baseline and a 93\% increase in AP$_{50}$. Even the Transformer-based RT-DETR-L~\cite{zhao2024detrs}, despite its large capacity and extensive pretraining, benefits from a 6.5\% mAP improvement with YCDa, highlighting the generality of our approach.

On the NC4K-D and CAMO-D datasets, all CNN-based models exhibit consistent accuracy improvements. YCDa-YOLO12s achieves 33.7\% mAP on NC4K-D (+29.6\%) and 26.0\% on CAMO-D (+27.5\%), validating the robustness of YCDa across datasets with varying scales and scene complexities. However, RT-DETR-L + YCDa shows slight performance degradation on these two datasets (34.1\% on NC4K-D and 22.5\% on CAMO-D). We attribute this to dataset-scale limitations: when training samples are limited (NC4K-D has only 48\% and CAMO-D 12\% the size of COD10K-D), the chrominance–luminance decoupling alters the feature distribution, requiring sufficient samples for large-capacity Transformers to re-adapt. CNN-based models, with stronger inductive biases, adapt more effectively to YCDa’s feature refinement, showing greater resilience under few-shot conditions. Overall, these results confirm that YCDa enables real-time CNN detectors to better distinguish discriminative texture cues from color noise, substantially improving detection accuracy with minimal computational cost.

\subsubsection{Qualitative Analysis.}  
Figure~\ref{fig:visualization} compares attention heatmaps between the baseline YOLO12s and YCDa-YOLO12s across the COD-D datasets. YCDa-YOLO12s demonstrates superior focus on the true camouflaged object regions, effectively separating targets from complex backgrounds. In challenging scenes—such as when the object is partially occluded by leaves (final row)—YCDa maintains consistent attention across the occluded areas, while the baseline primarily attends to high-saliency regions like the object’s head. This observation illustrates that YCDa encourages the model to rely on texture and contour cues rather than solely on color, leading to fewer false positives, missed detections, and re-identification errors. These qualitative results further validate the superiority of our bio-inspired decoupling and dynamic attention strategy.

\subsubsection{Efficiency Analysis.}  

To further evaluate the practical applicability of our YCDa strategy in real-world scenarios, we conduct a detailed runtime efficiency analysis using YOLO12s~\cite{tian2025yolov12} as the baseline on the COD10K-D dataset. As reported in Table~\ref{tab:realtime}, YCDa-YOLO12s achieves substantial accuracy gains while preserving real-time inference capability. Specifically, it attains 131.4 FPS on a single NVIDIA A100 GPU, only a 4.9\% reduction compared to the original baseline, despite integrating additional color decoupling and attention modules. This negligible overhead demonstrates that YCDa introduces no significant computational burden or structural redundancy.

Moreover, the model achieves an absolute improvement of +9.5 mAP and a relative increase of 112\%, indicating that YCDa effectively enhances camouflaged object perception without sacrificing speed—an essential requirement for time-sensitive tasks such as autonomous driving, robotic vision, and wildlife monitoring. These results collectively confirm that YCDa is both accurate and efficient, offering a lightweight yet powerful plug-in strategy that seamlessly integrates into existing real-time detectors.

\subsection{Ablation Studies}
We further analyze the contribution of each YCDa component through ablation studies on the COD-D benchmarks using YOLO12s as the baseline. As summarized in Table~\ref{tab:ablation}, both YCbCr conversion (YC) and Information-aware Channel Attention (ICA) independently yield substantial gains. On COD10K-D, YC alone raises mAP from 8.5\% to 16.3\% (+91.8\%), while ICA achieves a similar +95.3\% improvement, confirming the importance of chrominance–luminance separation and adaptive attention reweighting. When combined, YC + ICA further boosts mAP to 18.3\% (+115\%), showing strong synergy between the two modules.

The complete YCDa configuration (YC + ES + ICA) delivers additional improvements in NC4K-D and CAMO-D, reaching 33.7\% and 26.0\% mAP respectively. The ESSamp module, designed to replace pointwise convolutions, refines the decoupled features while maintaining channel independence, allowing richer chrominance–luminance information to flow into ICA. This mechanism proves particularly effective on medium-scale datasets, where refined channel-wise representations enhance ICA’s sensitivity to saliency variations. On the larger COD10K-D dataset, which contains many background-only samples, ESSamp may introduce mild feature noise, slightly limiting further improvement.

To further validate ICA’s effectiveness, we compare it with several mainstream attention mechanisms. As shown in Table~\ref{tab:attention_comparison}, ICA consistently outperforms others across all datasets. While SE~\cite{hu2018squeeze} and varOnly exhibit dataset-dependent strengths—SE favoring large, noisy datasets (COD10K-D) and varOnly performing better on texture-rich ones (NC4K-D, CAMO-D)—ICA integrates both global average and variance information, achieving the best of both worlds. It attains 18.1\% mAP on COD10K-D and 33.7\% / 26.0\% on NC4K-D / CAMO-D, outperforming CBAM and FCA by a clear margin. These results confirm that ICA’s dual-statistics design effectively balances robustness and sensitivity, enabling adaptive attention allocation across chrominance–luminance channels.

\section{Conclusion and Future Work}

\subsection{Conclusion}
This work addresses the challenges of scarce discriminative cues and strong noise in realistic camouflaged object detection (RCOD). Inspired by the biological visual system, we propose \textbf{YCDa}, an efficient early-stage feature processing strategy. YCDa decouples chrominance–luminance information via color space transformation, preserves channel independence through pointwise-convolution-free downsampling, and adaptively reallocates attention using an Information-aware Channel Attention module. Extensive experiments show that YCDa consistently boosts performance across various real-time detectors and establishes new state-of-the-art results on COD10K-D, NC4K-D, and CAMO-D with negligible computational cost.

\subsection{Future Work}
Future research will explore extending YCDa to Transformer-based architectures and temporal domains for video COD, where motion and illumination dynamics are crucial. Another direction involves learning adaptive color space transformations that enable the model to automatically discover task-specific spectral representations. Overall, YCDa provides an interpretable and lightweight framework that bridges biological vision mechanisms and modern detection models, offering new insights for robust perception of low-saliency targets in real-world environments.

\clearpage
{
    \small
    \bibliographystyle{ieeenat_fullname}
    \bibliography{ref}
}


\end{document}